\documentclass[10pt]{cai}

\graphicspath{{figs/}}

\begin{document}

\begin{center}

  \title{Balancing Information with Observation Costs in Deep Reinforcement Learning}
  \maketitle


  \begin{tabular}{cc}
    Colin Bellinger\upstairs{\affilone,*}, Andriy Drozdyuk\upstairs{\affilone}, Mark Crowley\upstairs{\affilthree}, Isaac Tamblyn\upstairs{\affilone,**}
  \\[0.25ex]
  {\small \upstairs{\affilone} National Research Council of Canada
Ottawa, Canada} \\
  {\small \upstairs{\affilthree} University of Waterloo, Waterloo, Canada} \\
  {\small \upstairs{**} Vector Institute for Artificial Intelligence, Toronto, Canada} \\
  \end{tabular}
  
  \emails{
    \upstairs{*}colin.bellinger@nrc-cnrc.gc.ca \upstairs{**} isaac.tamblyn@nrc-cnrc.gc.ca 
    }
  \vspace*{0.2in}
\end{center}

\begin{abstract}
The use of reinforcement learning (RL) in scientific applications, such as materials design and automated chemistry, is increasing. A major challenge, however, lies in fact that measuring the state of the system is often costly and time consuming in scientific applications, whereas policy learning with RL requires a measurement after each time step. In this work, we make the measurement costs explicit in the form of a costed reward and propose the active-measure with costs framework that enables off-the-shelf deep RL algorithms to learn a policy for both selecting actions and determining whether or not to measure the state of the system at each time step. In this way, the agents learn to balance the need for information with the cost of information. Our results show that when trained under this regime, the Dueling DQN and PPO agents can learn optimal action policies whilst making up to 50\% fewer state measurements, and recurrent neural networks can produce a greater than 50\% reduction in measurements. We postulate the these reduction can help to lower the barrier to applying RL to real-world scientific applications.\\
\end{abstract}

\copyrightnotice

\section{Introduction}

Deep reinforcement learning (DRL) has recently demonstrated great potential in challenging simulated sequential decision making environment \cite{mnih2015human,silver2017mastering}. Policy learning with reinforcement learning (RL) and DRL, however, requires long training times with significant exploration of the relationships between states, actions and rewards. Moreover, standard RL frameworks assume that a measurement of the current state of the environment is available at each time step without time delays nor explicit monetary cost. Therefore, even after DRL agents are develop and deployed, they can continue place a high state measurement burden on the system. Our work focuses on DRL for in scientific applications, such as materials design and automated chemistry, where measuring the state of the system is often costly and time consuming.

In materials design, for example, determining the microscopic state of the environment after applying a process (such as heat, stirring, \textit{etc.}), requires the use of costly measurement and characterization equipment \cite{chemputer2019, RobotScientist2020}. The same is true for problems of combinatorial optimisation via quantum-annealing \cite{COOL2020}, and sequential decision making in healthcare \cite{nam2021reinforcement}. As a result, the scientist will often select a sequence of processes to applying by utilizing their intuition, partial information or past experience of the resulting state trajectory. In such circumstances, more experienced scientists will have a better ability to infer outcomes, and thus, apply more processes without measuring the state of the environment. This enable them to expedite the experimental process and reduce cost. 

In scientific applications of this nature, the optimal agent will achieve the design goal and limit the total costs associated with carrying out the design. We denote the learning objective as maximizing the costed return, which is discounted rewards incurred whilst carrying out the process minus the costs incurred by the measurement policy. Although it is highly relevant, this sort of behaviour is largely missing in RL literature. The only exceptions to this are the recent works in \cite{bellinger2021active,nam2021reinforcement}. Unlike the previous work, however, here we focus on the development of a framework to enable the use of off-the-shelf DRL algorithms. We focus these because there exists a robust set of easy to use packages available to the community. As a result, they are likely be preferred by scientific users. 


To facilitate DRL agents to learn scientist-like behaviour of dynamically applying sequences processes without necessarily measuring the state of the system, we modify the standard RL framework in four ways:
    \begin{itemize}
        \item Incorporate an explicit measurement costs;
	    \item Expand the action-space to action tuples in which the tuple specifies the process to be applied to the system (such as heat, stirring, \textit{etc.}) and whether or not to measure the next state of the system;
	    \item Add a memory of the last measured state of the system to the environment. When the agent opts not to measure the next state, the last measured state is returned as the observation; and,
	    \item Add a stale observation flag (0 or 1) to the state representation. When the observation return by the environment is the current state of the system, 1 is appended to the end of the observation, otherwise 0 is appended. 
    \end{itemize}
As in the previous work \cite{bellinger2021active}, we refer to this class of problem as an \emph{active-measure Markov Decision Process (MDP) with explicit costs}. We use the term active-measure, because the agent decides if and when to measure the next state of the system. When the agent measures the next state, the observation returned to the agent is a fully observable representation of the true underlying state of the system, and a domain-specific measurement cost is deducted from the reward.

Active-measure MDPs with explicit costs are related to traditional partially observable MDP (POMDP) problems \cite{kaelbling1998planning} where the challenges of noisy, costly and missing state observation have previously been studied. Whereas in traditional POMDPs the agent cannot pay to remove the underlying uncertainty, in our setup the agent explicitly determines to see the next state of the system at a cost (fully observable) or see the last measure state of the system (partially observable) for free. Thus, this class problem resides in the middle ground between MDPs and POMDPs. Active-measure MDPs with explicit costs are significantly easier problem than classic POMDPs because of omnipresent option to access the fully observable state and the fact that the agent does not require a dynamics model to be learned or provided \textit{a-priori} \cite{papadimitriou1987complexity}.

Here, we implement the active-measure with explicit cost framework as a wrapper class for the Open AI gym \cite{brockman2016openai} which is available here: \url{https://github.com/cbellinger27/active-measureGymWrapper}. The wrapper class expands the state- and action-spaces, and incorporates the user-specified measurement costs thereby converting the environments in the Open AI gym to active-measure MDPs with explicit costs. We evaluate the DRL algorithms proximal policy optimization (PPO) \cite{schulman2017proximal}, dueling deep Q networks \cite{wang2016dueling} (DQN) and deep Q networks with recurrent memory (DRQN) \cite{hausknecht2015deep} on the active-measure  with explicit costs framework with the classic control environments Cartpole, Acrobot, and Lundar Lander.

We find that the DRL agents learn measurement policies that reduce the measurement costs whilst learning an action policies that efficiently achieve the goal. In particular, our results show that the modified framework enables DRL agents to learn to make up to 50\% fewer state measurements. Moreover, by extending the standard DRL architectures to incorporating recurrent neural networks, the agents can learn to reduce the number of measurements by more than 50\%.

\section{Related Work}

This work has some relationship to previous work on \textit{active reinforcement learning} \cite{akrour2012april,krueger2016active,schulze2018active} that modified the action space. In particular, we utilised the previously proposed strategy of action pairs that encode a directive about how to move and whether to measure the state of the environment or utilise an oracle. These previous works, however, involved using a human experts to address the challenge of defining a complete reward signal. Alternatively, our work is focused on reducing the costs associate with measuring the state of the environment. Active perception is also related to our work in that the agent can actively decide to take steps to improve its information about the world \cite{gibson1966senses}. The key distinction with our work is that the agent in active perception improves its information via self-modification and self-evaluation, rather than directly requesting fully observable state information at a cost.

Our work has a connection with POMDPs in that when the environment returns the last measured state to the agent, the true state is only partially observable \cite{kaelbling1998planning}. Mixed observable MDPs (MOMDPs) \cite{ong2010planning,smith2012heuristic} are also related to work based the fact that they provide a blend of fully and partially observable measurements of the next state. Whereas in MOMDPs this mix of observability is an underlying property that is determined by the environment, in our work the agent controls the mixing. In both POMDPs and MOMDPs the learning agent does not have a direct mechanism to remove uncertainty due to partial observablity. Moreover, planning algorithms for POMDP and MOMDPs generally require learning a model of the underlying dynamics or having \textit{a-priori} knowledge of the model \cite{papadimitriou1987complexity}

A POMDP agents can only indirectly affect their uncertainty by improving the observation model and by choosing actions that change the environment. By providing fully observable state information on demand, but at a cost, off-the-shelf DRL algorithms can be applied to active-measure MDPs with explicit costs. In our setup, the uncertainty is a function of agent's choice to forgo measuring the state of the environment to lower its observation costs.  The agent can always opt to measure the state of the environment to remove its uncertain at a cost. Therefore, it can be seen as a sub-category of POMDP that can be efficiently solved using off-the-shelf DRL. The effectiveness of a tabular RL setup of this framework over using classical POMDP methods was demonstrated in previous work \cite{bellinger2021active}


\section{Problem Formulation}

This work focus on the classic RL setup which includes the environment as a tuple: $(S, A, P, S^\prime, R, \gamma)$. These are the standard components of an MDP, where $S$ is the state-space, $A$ is the action-space, $P(s^\prime | s, a)$ is the state transition probabilities, $R(s, a)$ is the reward function, and $\gamma \in [0,1]$ is a discount factor. $P$ and $R$ are not known by the agent. 

In this work, we focus on episodic environments with continuous states, $S \in \mathbf{R}^n$, discrete action sets $A = \{1,...,|A|\}$, and stationary state-transition dynamics. The standard RL objective is to learn a policy $\pi: s \rightarrow a$ that maximises the \emph{return}, which is defined as the discounted sum of rewards: $ R =  \sum^{\infty}_{t=0} \gamma^t R(s_t, a_t) .$

As introduced above, our objective is to facilitate off-the-shelf DRL algorithms maximise the costed return. As a result, we incorporate the explicit measurement costs into the return as: 
\begin{equation}
    R = \sum^{\infty}_{t=0} \gamma^t R(s_t, a_t) - C(m_t),
\end{equation}
where $C(m_t)$ is measurement costs at time $t$. 

For DRL to learn a non-trivial policy that maximises the costed return, it needs the ability to explicitly decide which actions to take and whether or not to measure the next state at each time step. Formally, the agent learns an action and measurement policy $\pi: s \rightarrow a, m$. To enable DRL to achieve this behaviour, we propose a five simple modifications to the classic RL framework. 

The first modification is to expand the action space to action pairs (atomic-action, measurement directive). The atomic-action is the standard action (\textit{e.g.} add heat, move up, \textit{etc.}), and the measurement directive is a Boolean flag that indicates if the state of the environment should be measured and returned after the application of the atomic action.  

When the agent selects an action pair that gives the directive to measure the environment ($m=1$), the measurement is returned to the agent for use in the selection of the next action pair. In addition, we augment the environment to give it a memory of the last measured state. When the agent selects an action pair that gives the directive not to measure the environment ($m=0$), the environment returns the last measured state to the agent (\textit{i.e.,} a stale state observation). In order to learn a useful policy, however, the agent needs to know when it is seeing a current measurement of the system and when it is seeing a stale one. We provide for this by expanding the state space to include a Boolean flag at the end of the state representation. When the environment returns a stale state observation to the agent, the flag is set to zero, otherwise, it is set to one. As an alternative, this augmentation can be omitted if the agent is given memory of its last action pair. However, augmenting the state space rather than the agent enables the use of off-the-self DRL. The results are expected to be the same either way.

Most importantly, in order to encourage the agent to learn a policy that only measures the state of the environment when it is necessary, we add an explicit user-defined measurement cost to the environment. The environment subtracts this cost from the reward if and only if the agent uses the measure directive. Therefore, an agent which measures at every time step will also know the current state of the system, but the reward at each time step will be reduced. On the other hand, a agent that never measures will save the measurement costs, but also have a growing uncertainty about the underlying state of the system. A non-trivial policy will balanced the cost of information with the need for information in order to efficiently (in terms of time and cost) achieve goal.  

The our wrapper class for the Open AI gym \cite{brockman2016openai} can be accessed here: \url{https://github.com/cbellinger27/active-measureGymWrapper}. 

\section{Experimental Setup}

We evaluate the performance of PPO \cite{schulman2017proximal} and Dueling DQN \cite{wang2016dueling} on the costed version of the Open AI gym environments Cartpole, Acrobot and Lunar lander \cite{brockman2016openai}. The DRL algorithms are implemented using Pytorch and the Tianshou deep learning packages \cite{weng2021tianshou}, and the results were record with Weights and Biases \cite{wandb}. The experiments were executed on National Research Council of Canada's cluster consisting of 36 NVIDIA V100 GPU nodes.

In our experiments, we first compare the performance of the agents on the standard Open AI gym environments (we refer to this as the vanilla environments, vanilla=1) to the active-measure with costs version of the environments (vanilla=0). These results serve to reveal the impact (if any) of the active-measure with explicit costs wrapper class on policy learning in the target environments. The main focus of the subsequent experiments are: \textit{a}) an analysis of the agents ability to learn non-trivial policies in the active-measure with explicit costs environments as the measurement costs are increased, and \textit{b}) an analysis of the impact of costs on the agents learned measurement policy. All of the results reported below are averaged over 10 independent trials. 
\begin{figure*}[h]
\centering
    \includegraphics[scale=0.13]{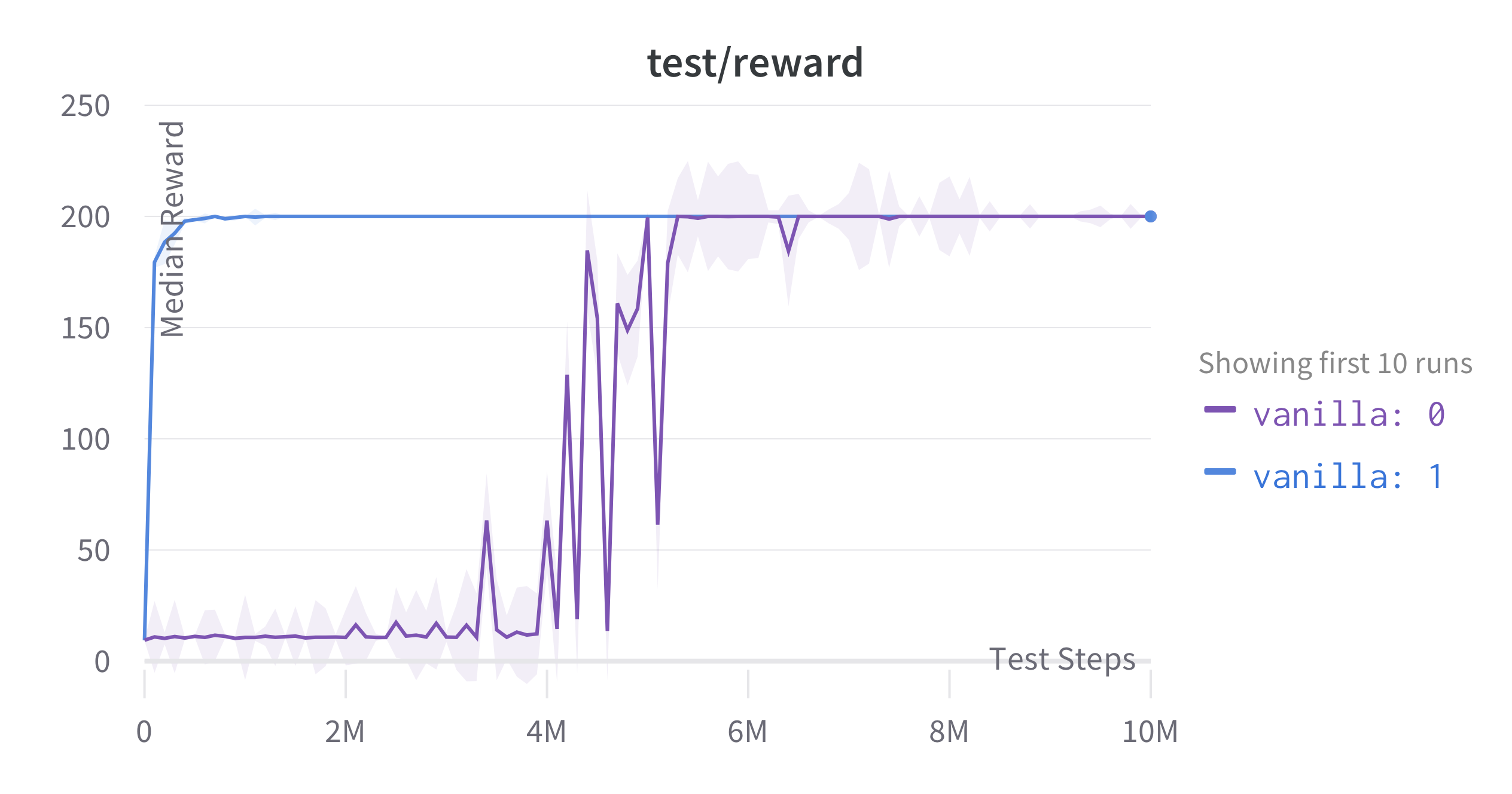}
    \includegraphics[scale=0.13]{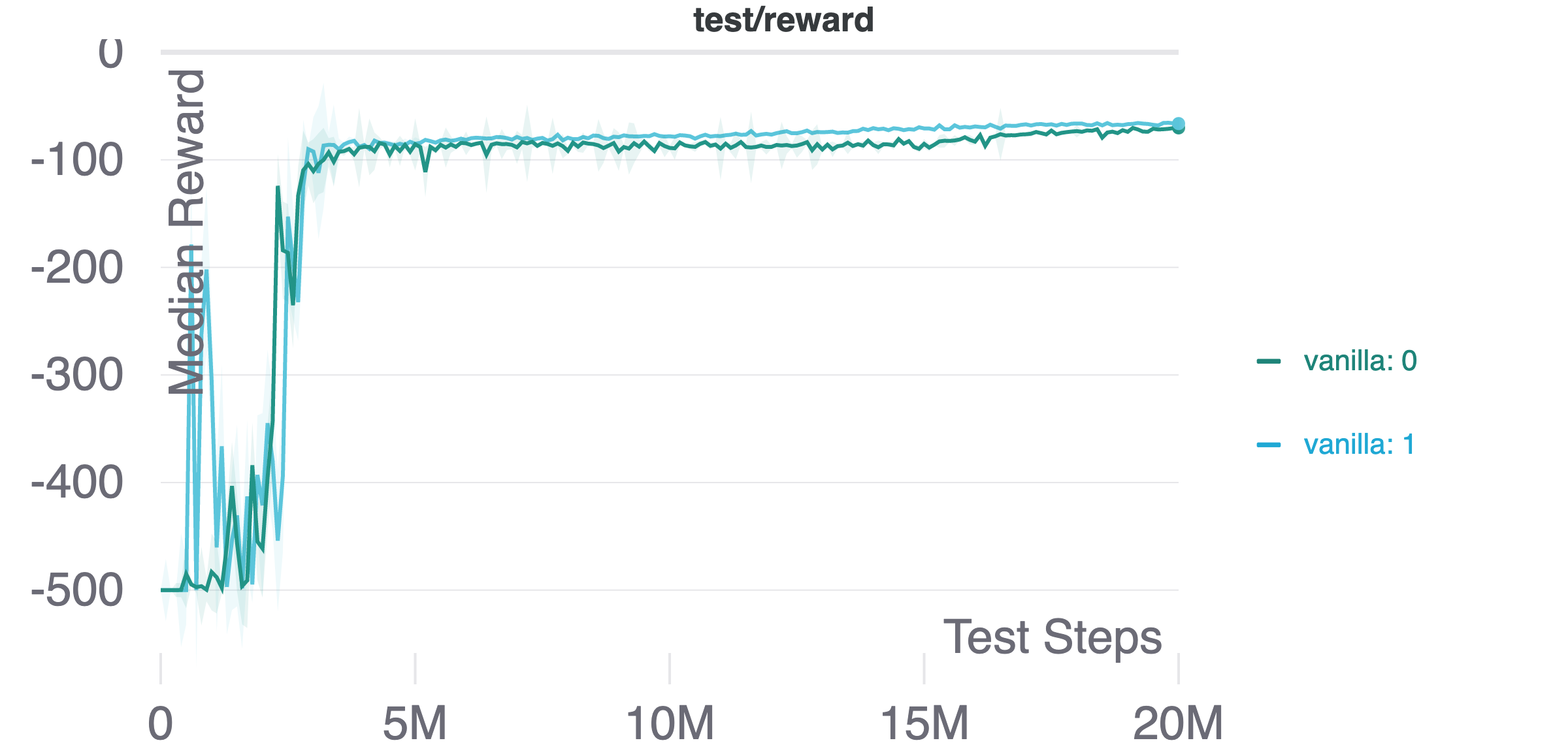}
\caption{Comparison of Dueling DQN on Cartpole (top) and Acrobot (bottom) environments in the vanilla mode and in the costed reward mode with cost equal 0. The results illustrate that expanding the state and action-space have minimal impact on policy learning.}
\label{fig:vanillAnalysis}
\end{figure*}

\section{Results}

\subsection{Vanilla Environment Versus Costed Environment}

This section evaluates if the active-measure with explicit costs wrapper has any strong negative impact on policy learning in the selected Open AI Gym environment. This analysis is conducted by comparing the agents' performance on the vanilla environments to the performance on the active-measure with explicit costs version of the environment with the measurement cost is set to 0. The corresponding results for Dueling DQN agents training on the Cartpole and Acrobot environments under both frameworks are presented in Figure \ref{fig:vanillAnalysis}. The results plot the median test rewards on the vanilla environment versus the costed reward version of environment with cost = 0. The results show the agents learning under the active-measure with explicit costs regime converge to the same median reward as on the vanilla environments.  This result also holds for Lunar Lander, but is omitted due to space considerations. 

We could expect some lag in the learning of agents on the active-measure with explicit costs versions of the environments because these have larger state and actions spaces. Indeed, we see this lag with agent learning on Cartpole, where it takes slightly longer to start learn, but not in the other environments. This is of interest because Cartpole is easiest of the three environment. More research is required to understand this phenomenon.

\subsection{Learning with a cost}

\begin{figure*}[h]
\centering
    \includegraphics[scale=0.07]{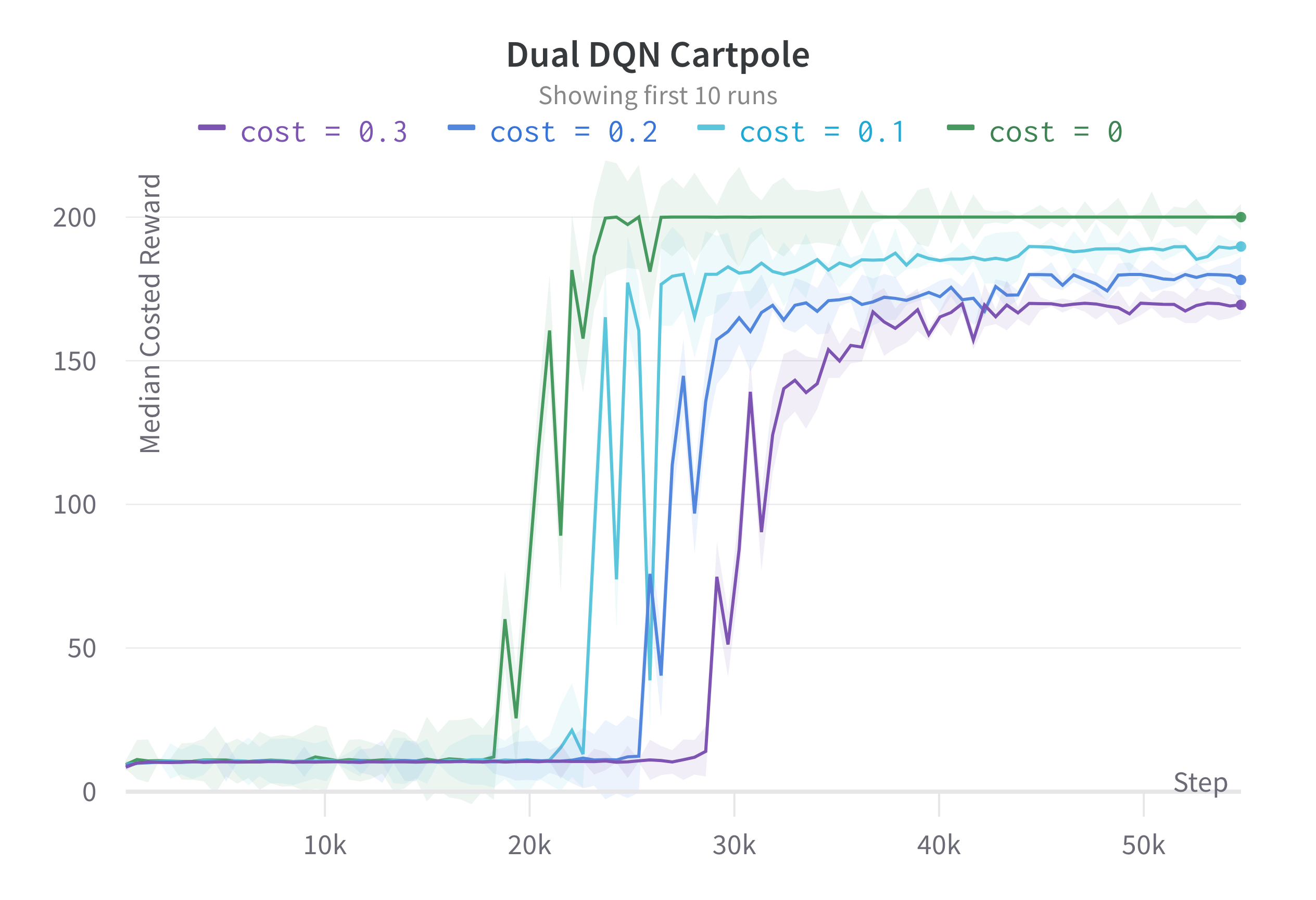}
    \includegraphics[scale=0.07]{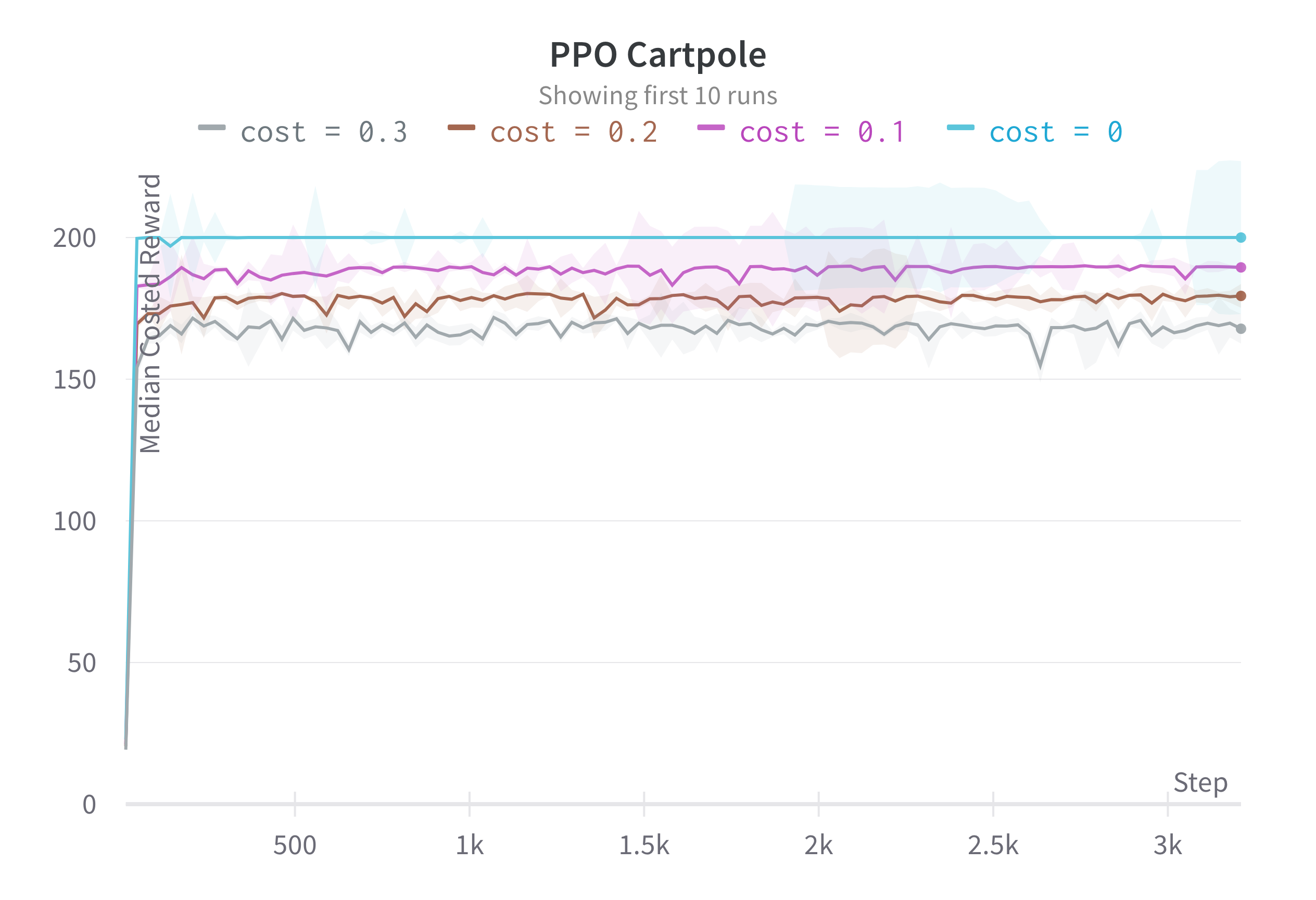}
    \includegraphics[scale=0.07]{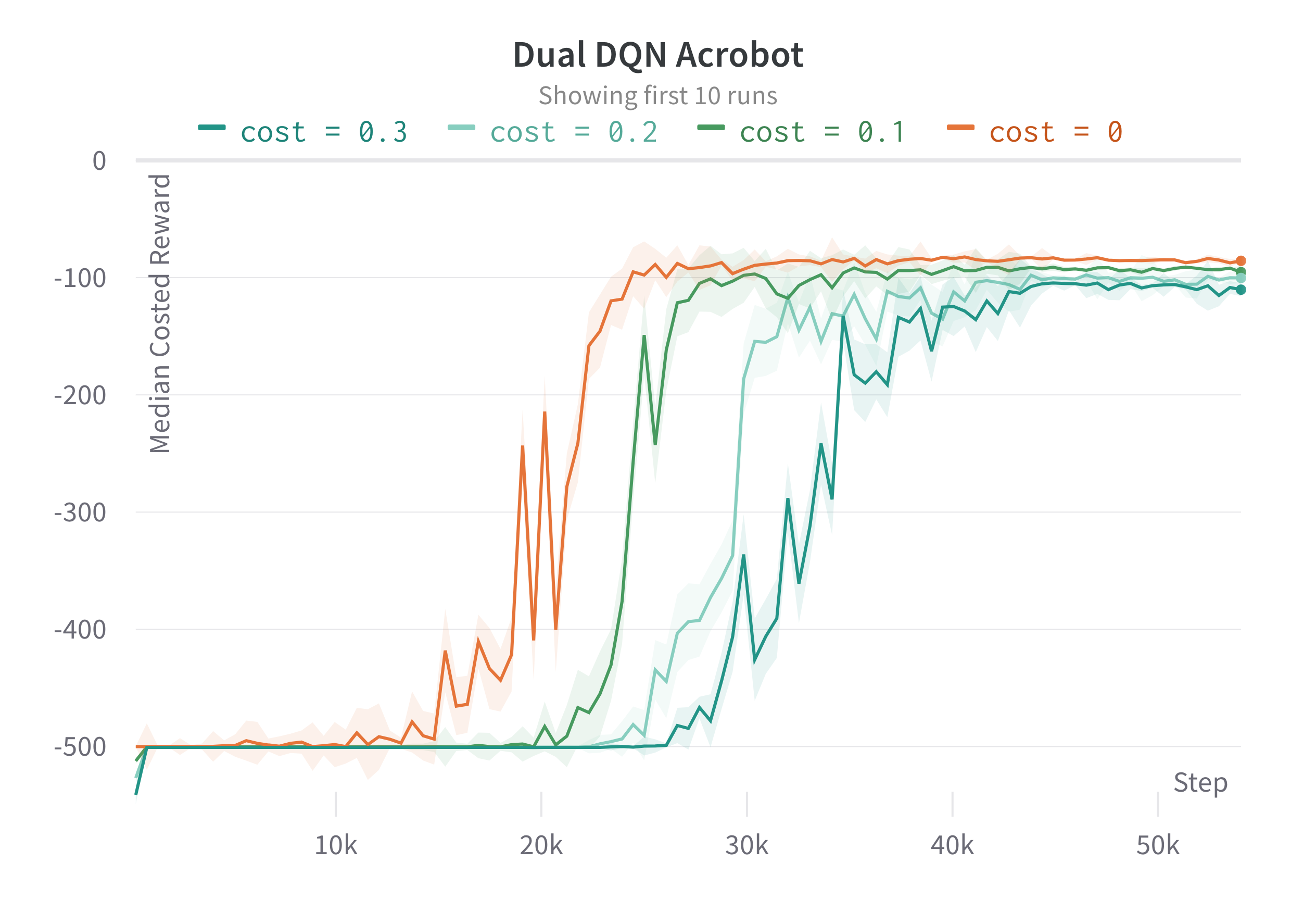}
    \includegraphics[scale=0.07]{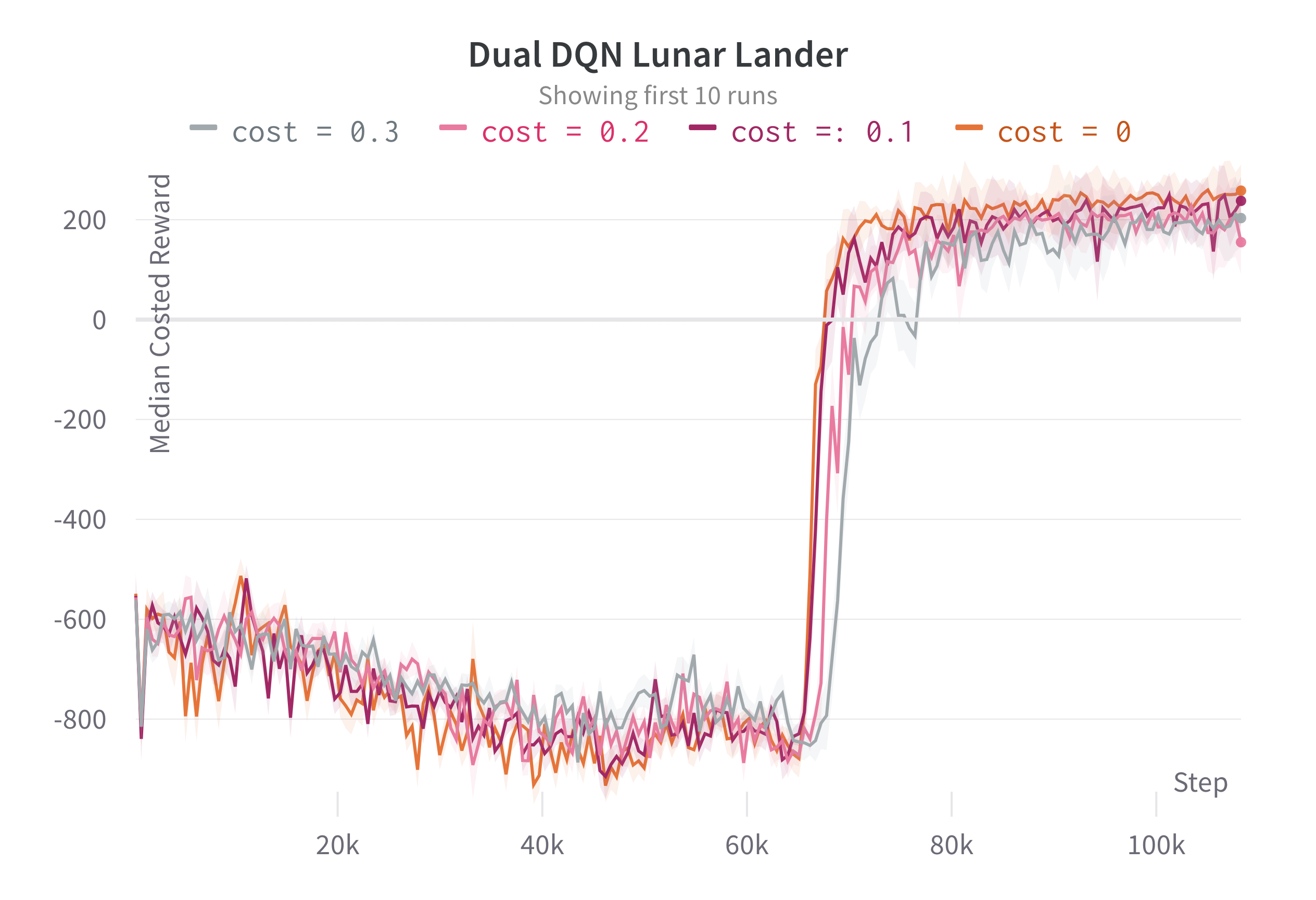}
\caption{Comparison of Dueling DQN Cartpole (top left), PPO Cartpole (top right), Dueling DQN Acrobot (bottom left) and Dueling DQN Lunar Lander (bottom right) in the active-measure with costs regime with increasing measurement costs. The results illustrate that increasing the cost reduces the costed rewards, but the agent is able to learn a policy that balances the frequency and cost of measurements with the need to obtain information to achieve the goal.}
\label{fig:CostAnalysis}
\end{figure*}

In this section, we evaluate the agents ability to learn in the active-measure with costs regime as the measurement costs are increased. The median costed reward results for Dueling DQN and PPO on Cartpole, Acrobot and Lunar Landerare depicted in Figure \ref{fig:CostAnalysis}  for costs 0, 0.1, 0.2 and 0.3. 

As verified in the previous section, the cost=0 curves show the optimal performance for the agents on each of the environments. Therefore, these serve as an ambitious, although unreachable, objective for the agents learning with costs greater than 0. Achieving this goal would require the agent to never measure the state of the environment, which can not be done in a model-free setting. Knowing this upper bound, however, aids our analysis. In the worst cases, the agent measures the state of the environment at every time step. This result in the agent's reward at each time step having the cost subtracted. Therefore, the lower bound for an agent with an optimal actions policy but measuring at each step is the maximum reward for the policy minus the measurement cost at each time step. For Cartpole, this bounds the costed reward of an agent that learns an optimal action policy between 200 and 200-(200*cost). Specifically, 180, 160 and 140 for costs of 0.1, 0.2, and 0.3 respectively. For Acrobot, this amounts to approximately -71, -78, and -84. We assess the agent's active-measure policies by how high above the lower bound converged costed reward is. 

The results for Cartpole and Acrobot show that whilst the agents' policies converge, the levels are well below the ambitious upper costed reward targets. Importantly, however, they are significantly above the baseline that solves the problem by measuring at every time step. On Cartpole, for example, the medians of the best costed rewards per trial are 190.03, 180.07, and 170.10. Therefore, the agents are able to learn non-trivial measurement policies that enable them to solve the problem and reduce their measurement costs. This performance trend holds for Lunar lander as well.   

\begin{figure*}[h]
\centering
    \includegraphics[scale=0.4]{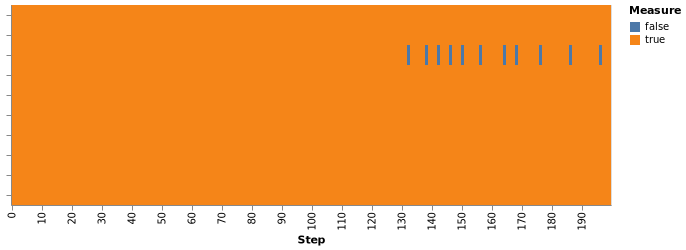}
    \includegraphics[scale=0.4]{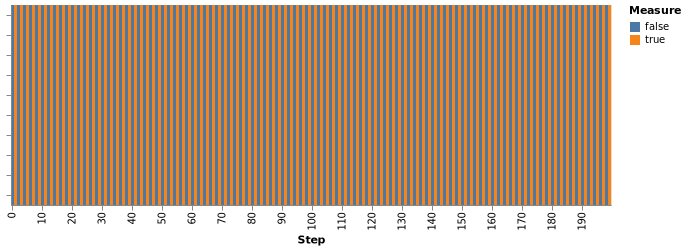}
\caption{Comparison of the measurement behaviour of Dueling DQN on Cartpole environments with cost of 0 (top) and 0.3 (bottom). The show that when the cost is greater than zero, the agent learns to measure only when necessary to achieve the goal.}
\label{fig:CartpoleMeasurementAnalysis}
\end{figure*}

\subsection{Learned Measurement policies}

In this section, we take a deeper dive into the measurement policies learned by the agents. Here, we aim to understand what the measurement policy behaviour looks like, and if / how it depends on measurement costs and the dynamics of the environment. 

To explore this question, we load the best policy learned by the respective agents and test them over 10 independent episodes. First, we compare the measurement policy learned on Cartpole with costs of 0 and 0.3. The plots in Figure \ref{fig:CartpoleMeasurementAnalysis} depict the measurement directives issued by the policy learned with Dueling DQN.  The \textit{x}-axis specifies the time steps of the episode and the \textit{y}-axis specifies the independent episode trial. At each time step, the colour of the rectangle indicates if the agent's learned policy requested a measurement or not. The first plot corresponds to a cost of 0 and illustrates that on the Cartpole environment with the cost set to 0, the agent learns to measure at each time step. This is clearly the best policy when measurements are free. 
\begin{figure*}[h]
\centering
    \includegraphics[scale=0.4]{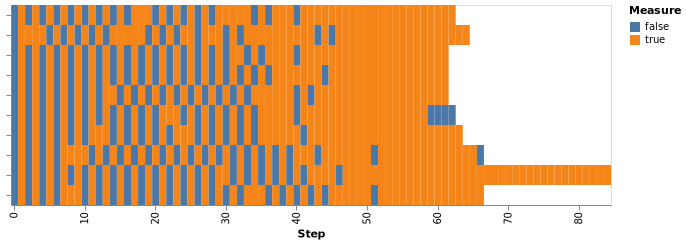}
    \includegraphics[scale=0.4]{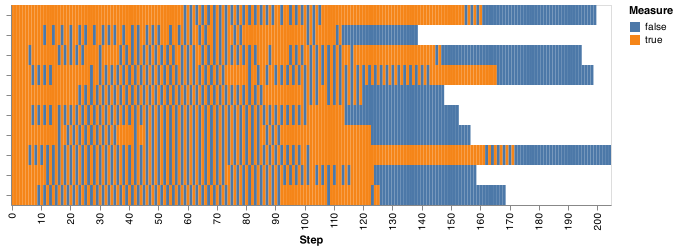}
\caption{Comparison of the measurement behaviour of Dueling DQN on Acrobot environments (top) and lunar lander environment (bottom) with cost of 0.2. This show that when the cost is greater than zero, the agent learns to measure only when necessary to achieve the goal. It also shown that the measurement policy is dependent upon the dynamics of the system.}
\label{fig:AcrobotMeasurementAnalysis}
\end{figure*}

In both the case of cost 0 and 0.3 the agents learn optimal action policies that enable the pole to stay upright until the maximum of 200 steps. The second plot in Figure \ref{fig:CartpoleMeasurementAnalysis} illustrates the measurement policy learned by Dueling DQN when the cost is set 0.3.   Here, we see the agent learns a policy that alternates between measuring and not measuring at consecutive time steps. This illustrates that when the measurement cost is set to be greater than zero, the agent learns to judicious measurement states in a manner that balances its need for state information with the costs of that information and the potential future reward.

The results for costs 0.1 and 0.2, which are withheld for space, show that Dueling DQN agent learns the same measurement policy for all costs greater than zero that are considered. This indicates that a standard DRL agent operating in the proposed costed setup can take at most one step without measuring the next state of the system. This a consequence of the design of the active-measure with explicit costs framework. To increase the number of steps without measuring in this setup more memory can be added to the environment, a model can be learned, or as we demonstrate below, the agent can be given recurrent memory. 

The results in Figure \ref{fig:AcrobotMeasurementAnalysis} demonstrate the measurement policies learned by Dueling DQN on the Acrobot and Lunar Lander environments with a measurement cost of 0.2. These results show that the agents learn more complex measurement policies for these environments than they learned for Cartpole. The patterns demonstrate that the measurement policies dependent on the dynamics of the environment. In the case of Lunar Lander (bottom), early in the episode, the policy takes measurement at each step (shown by the prevalence orange rectangles in the early part of the episode). We postulate that this is to account for the changes in environment that occur at the start of each episode. Once the agent has established an understanding of the environment in the current episode, the policy shifts to alternating between measuring and not measuring (alternating blue and orange rectangles in the middle of the episode.) Finally, when the lander is close to the landing  zone, the agent no longer needs to measure at all (all blue rectangles towards the end of the episode). This is where it is able to reduce the bulk of its state measurements. As a result, the agent is able to reduce its per episode measurement by nearly half. 

\begin{figure*}[h]
\centering
    \includegraphics[scale=0.4]{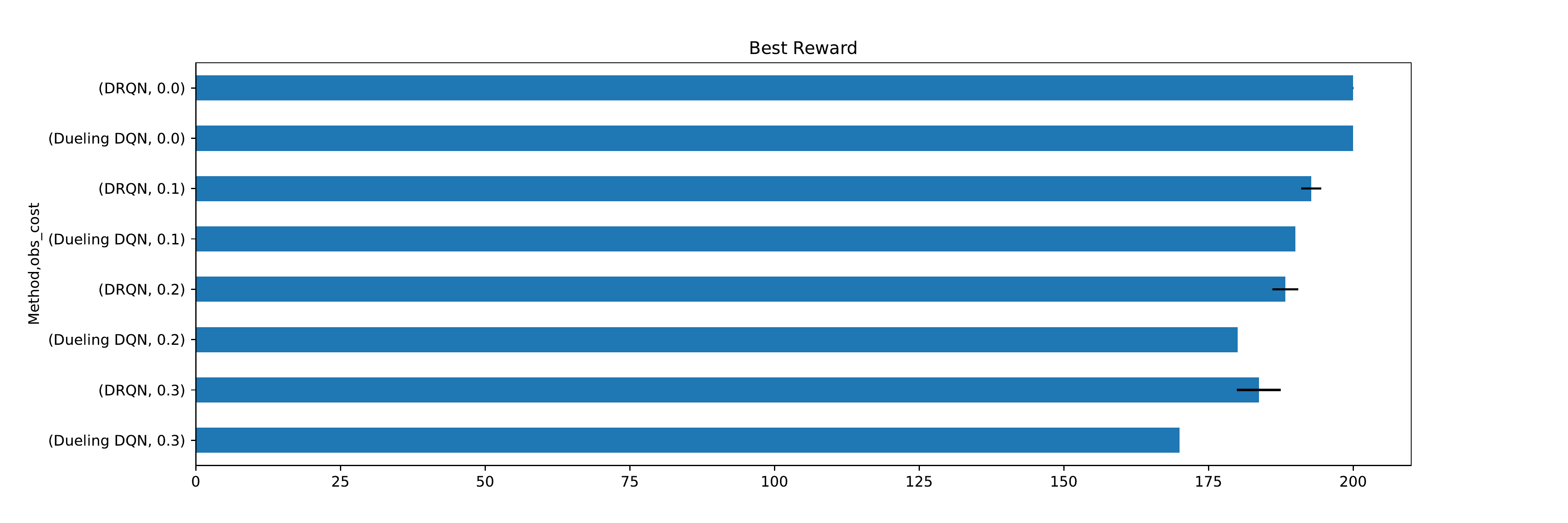}
\caption{Comparison of the best costed rewards of Dueling DQN and DRQN on Cartpole environments with increasing costs. As the costs increase, DRQN exploits is recurrent memory to measure less frequently and achive a higher costed reward.}
\label{fig:CartpoleMeasurementAnalysisBar}
\end{figure*}

For Acrobot, shown in the top plot of  Figure \ref{fig:AcrobotMeasurementAnalysis}, the agent learns to use the alternating measure-do-not-measure pattern for the first two-thirds of the episode. Acrobot, like Cartpole, starts each episode in the same position with the same dynamics. Therefore, the agent can forego some measurements at the start. As the agent gets closer to flipping the Acrobot over the goal line, the action choice becomes more sensitive and multiple state trajectories can lead to success. Thus, the agent learns that it needs to measure frequently as it gets closer to achieving the goal. The nature of the Acrobot dynamic significantly limit the agents ability to reduce its state measurements without impacting its learned action policy.

\subsection{Recurrent Memory}

The final set of results are presented in Figure \ref{fig:CartpoleMeasurementAnalysisBar} and Figure \ref{fig:CartpoleMeasurementAnalysis}. These results compare the use of Dueling DQN to DQN with a recurrent network (DRQN). The objective is to explore the hypothesis that adding recurrent memory to the agent will enable it to take consecutive steps without measuring the state of the environment. 

\begin{figure*}[h]
\centering
    \includegraphics[scale=0.4]{dqnCartpole_obs-cost_0_3experiment_0_0_98caead2807c45923ee1.png}
    \includegraphics[scale=0.4]{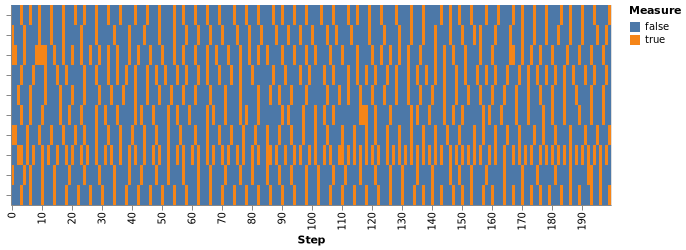}
\caption{Comparison of the measurement behaviour of Dueling DQN and DRQN on Cartpole environments with cost of 0.3. These plots show that DRQN is able to further reduce its measurement requirements under the active-measure with costs regime.}
\label{fig:CartpoleMeasurementAnalysis}
\end{figure*}

The results in Figure \ref{fig:CartpoleMeasurementAnalysisBar} compare the median of the best costed reward achieved by the  Dueling DQN and DRQN on Cartpole. The results are shown for costs of 0, 0.1, 0.2 and 0.3. For cost=0, both agents learn policies that achieve the optimal reward. As the cost is increased from 0, the plot shows that DRQN has a growing advantage in this environment in terms of the costed reward. It is also revealed that the costed reward of the Dueling DQN policy has a small standard deviation across test trails (the standard deviation bars are too small so see on this plot.) Conversely, the DRQN policy across trials has a noticeable standard deviation. Nonetheless, DRQN has a clear advantage. This is particularly the case with the largest measurement costs of 0.3. 

Figure \ref{fig:CartpoleMeasurementAnalysis} presents the measurement policies for  Dueling DQN and DRQN on Cartpole with the measurement cost of 0.3. In the case of Dueling DQN (top), we see the familiar alternating measurement strategy. Alternatively, in the DRQN plot (bottom), we see that the agent requires very few state measurements per episode to achieve the goal. This is depicted by the fact that the plot is dominated by blue rectangles with a few orange ones interspersed. Thus, DRQN learns an action policy that solves the environment whilst learning a measurement policy that takes consecutive steps without measuring the next state, thereby achieving a higher costed reward. This demonstrates that recurrent DRL can exploit its hidden state representation to track the true state of the environment within the costed reward regime. Importantly, it can utilise this to reduce the minimum measurement frequency required by the agent to achieve a higher costed reward.

\section{Discussion}


In laboratory science and design where measuring the state of the system is costly and time consuming, it is standard that the research workflow smoothly shifts between take real measurements of the state of the system and inferring sets of state based on textbook knowledge, past experience and expect trajectories. On the other hand, the standard RL framework requires a measurement at teach time step. Although RL is expected to be beneficial in many scientific applications, its need for a frequent and predetermined state measurement pattern can place a significant barrier on the application of RL to problems with high measurement costs. The above results demonstrate that the proposed costed reward regime enables off-the-shelf DRL algorithms to learn a non-trivial state measurement policies in a model-free manner whilst learning an optimal action selection policies. This is one step towards enable the use of RL in many scientific applications. 

In spite of the benefits demonstrated with the use of the proposed active-measure with costs regime, we see the reliance on action pairs as a limitation. The environments considered here had relatively small actions spaces. As a result, mapping the actions to action pairs, which doubles the size of the action space, had minimal negative impact on the effective learning rate. Real-world applications, however, often have a large number of discrete actions or have a continuous action space. The current solution does not have a straightforward mapping to continuous action spaces, a side from discretizing the action space, which is not always ideal.   

The results demonstrate DRL algorithms learning within the active-measure with costs regime can successfully learn a measurement policy that reduce the need for measurement. The best achievable reduction for a non-recurrent agent is to cut the measurement requirement in half. The potential to achieve such a reduction under the current framework, however, is limited by the complexity of the environment. In the case of Cartpole, which has simple dynamics and a deterministic start state, the maximum possible reduction is achieved. In more complex environments, the agent achieves meaningful reductions but can not fully reduce the measurements frequency to half of the standard framework. Although, both PPO and Dueling DQN can learn good action and measurement policy on the Cartpole environment, our analysis found that Dueling DQN was more robust on the more complex environments and to higher measurement costs. We postulate that this is because more exploration is required to find a measurement policy that will reduce the measurement cost below the naive baseline and still allow the learning of a good action policy. We surmise that the ease at which exploration can be tuned up or down in DQN to suit the environment makes it easier to learning a good policy. 

Our results for recurrent DRL demonstrate that it can help further reduce in the measurement frequency. However, we have also found that DRQN can suffer from high variance in learned policies, across random trails with a learned policy, and that this variance increase with the complexity of the target domain. The variance on the Acrobot and Lunar Lander environments had a significant impact on the average performance of the DRQN agent. More work and alternative architectures are need to improve upon this. Our on-going work is exploring the full potential of recurrent DRL to improve the measurement requirements in more complex applications. We expect that recent work such as \cite{zhu2017improving,kapturowski2018recurrent} may offer some new directions.

\section{Conclusion}
Many science-based applications have high temporal and monetary costs associated with measuring the state of the environment. This places a significant barrier on the use of RL in scientific applications such as materials design and pharmacology. As a result, developing strategies that reduce the number of state measurements demanded by RL is an important feature to improving its applicability. To achieve this, we postulate that the RL agent, much like a laboratory researcher, should be given the control to decided if and when it is necessary to measure the state of the system. From an RL perspective, this poses an interesting challenge that requires the agent to learn a policy that balances the need for information with the cost of obtaining it, while attempting to learn an optimal action policy.

This work proposed a set of modifications to the standard RL framework which results in the active-measure MDP with explicit costs framework. This new framework facilitates the use of off-the-shelf DRL in the active-measure regime. Our results show that within this setup, Dueling DQN and PPO can learn optimal action policies and a measurement policy that reduces the measurement cost. Using  DRL, the agents can learn to cut the total cost in up to half, and with the use of recurrent neural networks, agents can reduce the cost further by leveraging their internal memory to facilitate successive steps without measuring the state of the environment.  

Our on-going work is focused on developing a simulated chemistry lab environment as a test-bed that is closer to the real-world applications of interest. From a technical perspective, we are exploring  alternatives to the action pair setup so as to not require a doubling the action space, we are conducting more research into the use of recurrent neural network architectures, and we are exploring the incorporation of novel ideas from model-based RL and POMDP planning.  




\thispagestyle{plain}
\printbibliography


\end{document}